\documentclass{article}

\usepackage{mypackage}
\usepackage{natbib}
\setcitestyle{square, comma, numbers,sort&compress, super}
\usepackage{arxiv}
\usepackage{cleveref}
\usepackage[utf8]{inputenc} 
\usepackage[T1]{fontenc}    
\usepackage{url}            
\usepackage{booktabs}       
\usepackage{amsfonts}       
\usepackage{nicefrac}       
\usepackage{microtype}      
\usepackage{lipsum}
\usepackage{lineno}
\usepackage{caption}
\usepackage{mysymbols}
\usepackage{amsmath,amssymb, amsthm} 
\usepackage{stmaryrd}
\usepackage{latexsym}
\usepackage{xcolor}
\usepackage{bm}
\usepackage{appendix}
\usepackage{subfigure}
\usepackage[bottom]{footmisc}
\graphicspath{{figures/}}
\DeclareMathOperator{\E}{\mathbb{E}}
\DeclareMathOperator{\R}{\mathbb{R}}
\DeclareMathOperator{\A}{\mathbf{A}}

\DeclareMathOperator*{\argmax}{argmax}
\definecolor{green}{RGB}{94,193,101}

\title{Integrating independent and centralized multi-agent reinforcement learning for traffic signal network optimization}

\author{
  Zhi Zhang\thanks{Preprint. Work in progress},\hspace{5pt}  Jiachen Yang,\hspace{5pt}  
   Hongyuan Zha \\
  Georgia Institute of Technology\\
  \texttt{\{zhizhang, jiachen.yang\}@gatech.edu zha@cc.gatech.edu} \\
}

\begin{document}

\maketitle

\begin{abstract}
Traffic congestion in metropolitan areas is a world-wide problem that can be ameliorated by traffic lights that respond dynamically to real-time conditions.
Recent studies applying deep reinforcement learning (RL) to optimize single traffic lights have shown significant improvement over conventional control.
However, optimization of global traffic condition over a large road network fundamentally is a cooperative multi-agent control problem, for which single-agent RL is not suitable due to environment non-stationarity and infeasibility of optimizing over an exponential joint-action space.
Motivated by these challenges, we propose QCOMBO, a simple yet effective multi-agent reinforcement learning (MARL) algorithm that combines the advantages of independent and centralized learning.
We ensure scalability by selecting actions from individually optimized utility functions, which are shaped to maximize global performance via a novel consistency regularization loss between individual utility and a global action-value function.
Experiments on diverse road topologies and traffic flow conditions in the SUMO traffic simulator show competitive performance of QCOMBO versus recent state-of-the-art MARL algorithms.
We further show that policies trained on small sub-networks can effectively generalize to larger networks under different traffic flow conditions, providing empirical evidence for the suitability of MARL for intelligent traffic control.

\end{abstract}


\section{Introduction}
With increasing urbanization, traffic congestion is a significant and costly problem \cite{guerrini2014traffic,mcnew2014}.
While early works proposed to optimize traffic light controllers based on expert knowledge and traditional model-based planning \cite{porche1999adaptive,gershenson2004self,cools2013self}, there are promising recent results on applying flexible model-free methods in reinforcement learning (RL) \cite{sutton2018reinforcement} and deep RL, such as DQN in particular \cite{mnih2015human}, to find optimal policies for traffic light controllers that dynamically respond to real-time traffic conditions \cite{abdulhai2003reinforcement,genders2016using,li2016traffic,wei2018intellilight}.
These works model a single traffic light as a Markov decision process (MDP) equipped with a discrete action space (e.g. signal phase change) and a continuous state space (e.g. vehicle waiting time, queue length), and train a policy to optimize the expected return of an expert-designed reward function.

However, the single-agent RL perspective on traffic control optimization fails to account for the fundamental issue that optimizing global traffic flow over a densely connected road network is a cooperative multi-agent problem, where independently-learning agents face difficulty in finding global optimal solutions.
For example, if an intersection with low vehicle density in the North-South direction selfishly lets East-West traffic flow with little interruption to maximize its own performance, it will cause severe issues for any adjacent intersection that has heavy North-South traffic.
Instead, all traffic light agents must act cooperatively to optimize the global traffic condition while optimizing their own individual reward based on local observations.

On the other hand, existing work that adopt the multi-agent perspective on traffic signal optimization either fall back to independent learning \cite{liu2017cooperative} or resort to centralized optimization of coordinated agents \cite{arel2010reinforcement,van2016coordinated}.
Independent learners \cite{tan1993multi} only optimize their own reward based on local observations, cannot optimize for global criteria (e.g., different priorities for different intersections), and they face a nonstationary environment due to other learning agents, which violates stationarity assumptions of RL algorithms.
While centralized training can leverage global information, it requires maximization over a combinatorially-large joint action space and hence is difficult to scale.


Motivated by these challenges, our paper focuses on deep multi-agent reinforcement learning (MARL) for traffic signal control with the following specific contributions:

\begin{figure}[t]
\centering
\includegraphics[scale=0.7]{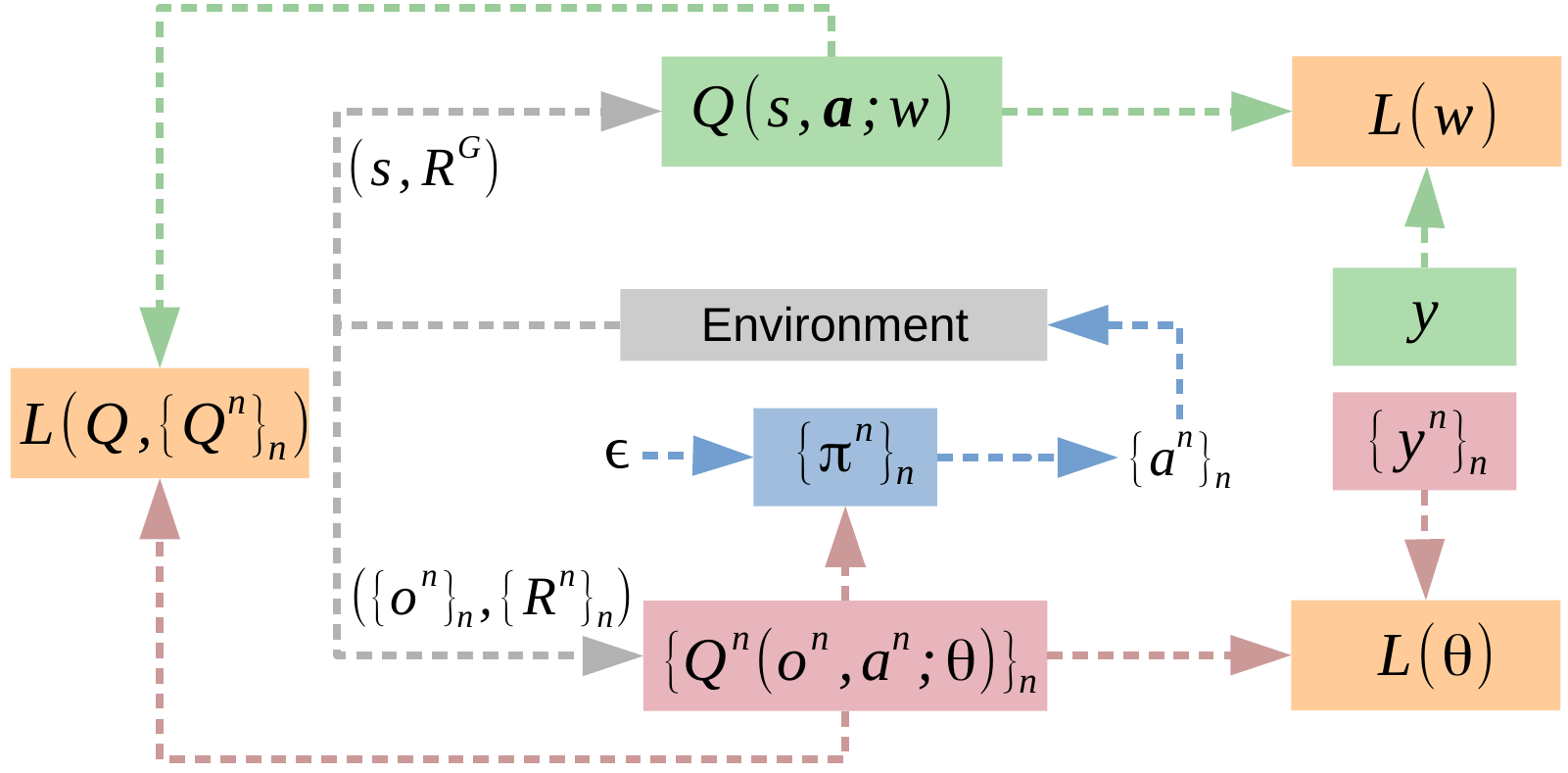}
\caption[one figure.]{QCOMBO architecture combining independent learning of $Q^n(o^n,a^n)$ with centralized training of $Q(s,\abf)$ via a novel consistency loss $L(Q,\lbrace Q^n \rbrace_n)$}
\label{Fig:QCOMBO}
\end{figure}

\textbf{1. Novel objective function combining independent and centralized training.} 
We propose QCOMBO, a Q-learning based method with a new objective function that combines the benefits of both independent and centralized learning (\Cref{Fig:QCOMBO}).
The key insight is to learn a global action-value function using the global reward, employ agent-specific observations and local rewards for fast independent learning of local utility functions, and enforce consistency between local and global functions via a novel regularizer.
Global information shapes the learning of local utility functions that are used for efficient action selection.

\textbf{2. Evaluation of state-of-the-art MARL algorithms on traffic signal optimization.}
Recent cooperative MARL algorithms
specifically tackle the case when all agents share a single global reward \cite{foerster2018counterfactual,sunehag2017value,rashid2018a}.
However, as these methods were not designed for settings with individual rewards, it is open as to whether their performance can be surpassed by leveraging such agent-specific information.
While they have shown promise on video game tasks, to the best of our knowledge they have not been tested on the important real-world problem of optimizing traffic signal over a network.
Hence we conducted extensive experiments comparing our algorithm versus independent Q-learning (IQL), independent actor-critic (IAC), COMA \cite{foerster2018counterfactual}, VDN \cite{sunehag2017value} and QMIX \cite{rashid2018a} on a variety of road networks with varying traffic conditions.

\textbf{3. Generalizability of traffic light control policies.}
To the best of our knowledge, we conduct the first investigation on the generalizability and transferability of deep MARL policies for traffic signal control.
Reinforcement learning methods are especially suitable for dynamic traffic light control since the transfer of a policy learned in simulation to real-world execution is arguably more feasible than in other domains (e.g. robotics).
Similar to domains where deep RL excels \cite{mnih2015human}, each traffic light has a small set of discrete actions for a signal phase change, which poses negligible issues for sim-to-real transfer.
Given improvements in sensor technology, measurements of traffic conditions can be increasingly accurate and real-world measurements can approach ideal simulated data.
Powerful model-free RL methods also do not require an accurate transition model that predicts traffic flow.
Hence, there is strong motivation to investigate whether a decentralized policy trained with simulated traffic approximating real-world conditions can be transferred to larger networks and different traffic conditions without loss of performance.

\section{Related Work}

Early work demonstrated the application of RL to single traffic light control \cite{abdulhai2003reinforcement}.
The success of deep RL has spurred recent works that incorporate high dimensional state information into a more realistic problem definition \cite{genders2016using,mousavi2017traffic,wei2018intellilight,liang2018deep},
which we further extend
to define the observation and action spaces of our new multi-agent setting.
Various choices of the reward function were proposed for training a single traffic light agent
\cite{balaji2010urban,chin2011q,mousavi2017traffic}.
We extended the definition of a single-agent reward \cite{wei2018intellilight}
by defining the global reward as a weighted sum of individual rewards using the PageRank algorithm \cite{page1999pagerank}.

Previous work on multi-agent traffic light control mostly relied on independent Q-learning (IQL) with heuristics to account for nonstationarity and coordination, such as:
single-agent Q-learning for a central intersection surrounded by non-learning agents \cite{arel2010reinforcement};
applying a Q function learned on a sub-problem to the full problem with max-plus action-selection \cite{van2016coordinated};
training only one agent during each episode while fixing other agents' policies \cite{liu2017cooperative};
sharing information among neighboring Q-learning agents \cite{el2013multiagent,liu2017intelligent}.
These approaches do not account for the importance of macroscopic measures of traffic flow \cite{geroliminis2007macroscopic}.
In contrast, our formulation explicitly shapes the learning of individual agents via a global reward.

Recent work proposed more sophisticated deep MARL algorithms for cooperative multi-agent problems with a global reward \cite{sunehag2017value,foerster2018counterfactual,rashid2018a}, under the paradigm of centralized training with decentralized execution \cite{bernstein2002complexity}.
However, these methods only learn from a global reward without using available individual rewards,
which motivates our proposal for a simple yet effective way to combine individual and centralized training.
To the best of our knowledge, these algorithms have yet to be evaluated and compared in the real-world problem of multi-agent traffic light control, which we do as part of our main contributions.

\section{Markov game model of multi-agent traffic light control}

We formulate the multi-agent traffic light control problem as a partially-observed Markov game $\langle S, \lbrace O \rbrace^n, \lbrace A \rbrace^n, P, R, N, \gamma \rangle$, consisting of $N$ agents labeled by $n = [1..N]$, defined as follows:

\textbf{Agents}
$n \in [1..N]$. 
Each agent controls the phase of one traffic light at an intersection.

\textbf{Observation space}
$O^n$.
Since all traffic lights have the same measurement capabilities, all agents' observation \textit{spaces} $O := O^1 = \dotsm = O^N$ have the same definition.
Each agent's individual observation \textit{vector} $o^n \in O$ depends on its own local traffic flow,
with the following components: $q^n\in\R^l$, $v^n\in\R^l$, $wt^n\in\R^l$, $delay^n\in\R^l$ (for $l$ incoming lanes at a traffic light), $ph^n\in\R^2$, and $d^n\in\R$, defined as:
\begin{itemize}[leftmargin=*]
    \item $q^n$: the length of queue on incoming lanes, defined as the total number of halting vehicles (speed less than 0.1m/s); 
    \item $v^n$, the number of vehicles on each incoming lane;
    \item $wt^n$, the average waiting time of all vehicles on each incoming lane; defined as the time in minutes a vehicle spent with a speed below 0.1m/s since the last time it was faster than 0.1m/s
    \item $delay^n$, the average delay of all vehicles on each incoming lane, the delay of a lane is equal to 1 - (average vehicle speed)/(maximum allowed vehicle speed);
    \item $ph^n$: the traffic light's current phase, indicating the status of the east-west and north-south directions, represented by a one-hot variable $ph^n: EW\times NS\mapsto \lbrace 0, 1 \rbrace^2$;
    \item $d^n$: phase duration in seconds since the last phase change.
\end{itemize}
Prior work used an image representation of positions of all vehicles near a traffic light and required convolutional networks \citet{wei2018intellilight}.
In contrast, we show this is not necessary and hence significantly reduce computational cost. 

\textbf{Global state space} 
$S$ contains all global information including every route and traffic light.
Global state $s \in S$ is the concatenation of all local observation vectors.

\textbf{Action space}
$A^n$.
Traffic controllers with the same capabilities means $A := A^1 = \dotsm = A^N$.
Extension to controllers with different types is easily done by learning separate $Q$ functions for each type.
The action $a^n$ of each agent is a binary variable to indicate whether or not the traffic light will keep the current phase or switch to another phase.
This definition is sufficient for policy learning because the agent's current phase is included in its own observation vector.
The game has joint action space $\A \equiv A^1\times...\times A^n$.
Agents produce joint action $\abf := (a^1,\dotsc,a^N) \in \A$ at each time step.
Let $\abf^{-n}$ denote all actions \textit{except} that of agent $n$.

\textbf{Individual reward}
$R^n(s,\abf) : S \times \A \mapsto \R$.
We base our individual reward on previous work that used weighted route features as the reward for the single-agent traffic light setting \cite{van2016coordinated,wei2018intellilight}
There are seven features used to calculate the individual reward (\Cref{app:reward-definition}).
The individual reward is a combination of these meaningful features that capture many intuitive metrics of desirable and undesirable traffic conditions.

\textbf{Global reward} 
$R^g$, defined as a weighted sum of individual rewards.
We explored different methods to compute these weights.
We used the PageRank algorithm to compute weights on each individual reward \cite{page1999pagerank}, since traffic intersections with higher risk of congestion are generally located in the central areas of the map that have higher interactions with surrounding traffic, and therefore should receive higher priority.
Hence $R^g(s,\abf) := \sum_{n=1}^N k_n R^n(s,\abf)$, where $k_n = \text{PageRank}(n)$.
While we considered using the traffic flow conditions under a fixed control policy to compute the weights for each traffic light, this is not a good choice since 
an arbitrary suboptimal nominal policy may produce a bad estimation of weights.
In contrast, the PageRank algorithm accounts for the topological structure of the transportation network, addresses the connectivity and interaction between agents, and assigns higher weights to the rewards of highly-connected traffic lights. 

\textbf{Evaluation criteria.}
Given a reward function designed with sufficient expert domain knowledge and specified with enough precision to disambiguate different traffic states, we can investigate the performance of state-of-the-art MARL algorithms by directly evaluating them using the cumulative reward and components of the reward.
Hence we do not resort to manual inspection of policy behavior, in contrast to previous work where certain states were aliased (i.e. produce the same reward) and manual inspection of the policy was required \citet{wei2018intellilight}.

\section{Multi-agent Reinforcement Learning Algorithms}

In this section, we give an overview of early and recent MARL algorithms, focusing on their respective strengths and weaknesses, with details in \Cref{app:algs}.
We use them as baselines for our experiments.

\textbf{IQL and IAC.}
Independent Q-learning (IQL) and independent actor-critic (IAC) have demonstrated surprisingly strong performance in complex multi-agent systems \cite{tan1993multi,tampuu2017multiagent,foerster2018counterfactual,rashid2018a}.
For each agent $n$, IQL directly applies single-agent Q-learning to learn a local utility function $Q^n(o^n,a^n)$, which is not a true action-value function because the presence of other learning agents result in a nonstationary environment from any single agent's perspective.
Similarly, IAC directly applies the single-agent policy gradient \cite{sutton2000policy} to train an actor-critic pair for each agent, resulting in actors $\pi^n(a^n|o^n)$ and critics $Q^n(o^n,a^n)$.
While IQL and IAC agents display strong ability to optimize individual rewards \cite{tan1993multi,yang2018cm3}, the lack of global information and a mechanism for cooperation means they are likely to settle for sub-optimal solutions.

\textbf{COMA.}
In cooperative MARL with a single global reward, COMA \cite{foerster2018counterfactual} estimates a centralized action-value function $Q^{\pibf}(s,\abf)$ to compute a counterfactual advantage function for a multi-agent policy gradient.
Their formulation is a low variance gradient estimate, as the advantage function evaluates the contribution of an agent's chosen action $a^n$ versus the average of all possible counterfactuals $\hat{a}^n$, keeping other agents' $a^{-n}$ fixed.
However, since the only learning signal comes from the global reward and individual agents are not directly trained to improve local performance, COMA may exhibit slower training in cooperative traffic light control.

\textbf{VDN.}
While IQL agents cannot learn to cooperate for a global reward, it is also not feasible to learn a single optimal action-value function $Q^*(s,\abf)$ since the maximization step requires searching over $|\Acal|^N$ joint actions.
Instead, VDN \cite{sunehag2017value} learns a joint action-value function that decomposes as $Q^{\text{VDN}}(s,\abf) := \sum_{n=1}^N Q^n(o^n,a^n)$, so that agents act greedily with respect to their own utility functions, while global reward is used for overall training.
However, there is no guarantee in general settings that the true optimal $Q^*(s,\abf)$ can be decomposed as a linear combination of individually-optimized utilities, which could limit VDN's performance.

\textbf{QMIX.}
QMIX \cite{rashid2018a} generalizes VDN by representing the optimal action-value function as a nonlinear function $Q^*(s,\abf) = F(Q^1,\dotsc,Q^N)$ of individual utility functions, while ensuring that the combination of individual $\argmax$ on each $Q^n$ yields the same joint action as a global $\argmax$ on $Q^*(s,\abf)$.
This is achieved by enforcing positive weights in the nonlinear mixing network $F$.
Despite being a more expressive model than VDN, the stability of QMIX depends on appropriate choices of mixing network architecture, for which there is little theoretical guidance, and QMIX also relies on global reward without using local reward for training.

\section{Method}



We propose QCOMBO, a novel combination of centralized and independent learning with coupling achieved via a new consistency regularizer.
We optimize a composite objective consisting of three parts: an individual term based on the loss function of independent DQN, a global term for learning a global action-value function, and a shaping term that minimizes the difference between the weighted sum of individual Q values and the global Q value.
This algorithm ensures that agents cooperate to maximize the global reward, which is difficult for independent learning agents to achieve, and also maintain the ability to optimize their individual performance using agent-specific observations and rewards, which is more efficient than a purely centralized approach.


\subsection{Individual Part}
Using individual observations and rewards for each agent is computationally efficient, since local observations generally have lower dimension than global state information, and also algorithmically efficient, since it avoids the difficult credit assignment problem of decomposing a single global reward signal for each agent to learn.
Furthermore, by optimizing individual utility functions $Q^n$ instead of a global optimal Q function, we reduce the maximization problem at each step of Q-learning from $O(|\Acal|^N)$ to $O(N|\Acal|)$.
Parameterizing the local utility function for agent $n$ with parameter $\theta^n$, we minimize the loss
\begin{align}
\mathcal{L}(\theta^n) &= \frac{1}{N} \sum_{n=1}^N \E_{\pibf} \Bigl[ \frac{1}{2}(y^n_t - Q^n(o^n_{t}, a^n_t; \theta^n))^2 \Bigr] \label{eq:loss-Q-individual} \\
y^n_t &= r^n_t + \gamma \max_{\ahat^n}Q^n(o^n_t,\ahat^n,\theta'^n) \label{eq:td-target-individual}
\end{align}
Since the agent population is homogeneous (i.e. all agents have the same observation and action spaces), we improve memory and computational efficiency by employing \textit{parameter-sharing} \cite{foerster2018counterfactual} among all agents, which means $\theta := \theta^n, \forall n \in [1..N]$.
Agents still act differently since they receive different observations, and we further give an agent indicator as input for agent disambiguation.
$\hat{\theta}$ are parameters of a target network \citet{mnih2015human}.

\subsection{Global Part}
Without global information, independently learning agents face a nonstationary environment due to the presence of other learning agents, and they may have insufficient information to find cooperative optima.
On the other hand, training an optimal global Q function is not scalable, since the Q-learning step would require maximization over $|\Acal|^N$ possible joint actions for $N$ agents.
To address this dilemma, our key insight is that we can learn the global Q function \textit{under the joint policy induced by all agents' local utility functions}, rather than learn the \textit{optimal} global Q function, and use it to shape the learning of individual agents via information in global state $s$ and global reward $R^g$.
Specifically, the joint policy defined by $\abf \sim \pibf(\abf|s) = \lbrace \argmax_{a^n} Q^n(o^n,a^n) \rbrace_{n=1}^N$ is associated with a global action-value function (letting $R^g_t := R^g_t(s,\abf)$):
\begin{align}
    Q^{\pibf}(s,\abf) := \Ebb_{\pibf} \Bigl[ \sum_{t=0}^{\infty} \gamma^t R^g \mid s_0=s,\abf_0=\abf \Bigr]
\end{align}
Parameterizing $Q^{\pibf}_w(s,\abf)$ with $w$, we minimize the loss:
\begin{align}
    &\Lcal(w) = \Ebb_{\pibf} \Bigl[ \frac{1}{2}\bigl( y_t - Q^{\pibf}_w(s_t,\abf_t) \bigr)^2 \Bigr] \label{eq:loss-Q-global} \\
    &y_t = R^g_t + \gamma Q^{\pibf}_{\hat{w}}(s', \abf')\vert_{a'^n = \argmax_{a^n} Q^n_{\hat{\theta}}(o'^n,a^n)} \label{eq:td-target-global}
\end{align}
where we let $(\cdot)' := (\cdot)_{t+1}$.
Crucially, action selection for computing the TD target \eqref{eq:td-target-global} uses the greedy action from local utility functions and does not use the global Q function.
The collection of local utility functions induce a joint policy $\pibf$ that generates data for off-policy learning of the global action-value function $Q^{\pibf}$.
$\hat{w}$ are target network parameters.

\subsection{Combined objective}
If each agent greedily optimizes its own local utility function, the global return can be suboptimal.
For example, if agent $n$ (with low weight $k^n$) has no flow in the N-S direction while adjacent agent $m$ (with high weight $k^m$) has heavy flow in the N-S direction, the individual optimal policy for $n$ is to let W-E traffic flow continuously to $m$, which negatively impacts conditions at $m$ and leads to low global reward.
This is supported by experimental results in a 1x2 network.
To address the suboptimality of independent learning, we propose a new consistency regularization loss
\begin{align}\label{eq:regularizer}
    \Lcal_{reg} := \Ebb_{\pibf} \Bigl[ \frac{1}{2}\bigl( Q^{\pibf}_w(s,\abf) - \sum_{n=1}^N k^n Q^n_{\theta}(o^n,a^n) \bigr)^2 \Bigr]
\end{align}
between global $Q^{\pibf}_w$ and individual utility functions $Q^n_{\theta}$.
Since $Q^{\pibf}_w$ is the true global action-value function with respect to the induced joint policy, this regularization brings the weighted sum of individual utility functions closer to global expected return, so that the optimization of individual utility functions is influenced by the global objective rather than purely determined by local information.
Hence the regularizer prevents any individual agent from attaining high individual performance at the cost of collective performance.


The complete QCOMBO architecture (\Cref{Fig:QCOMBO}) combines the individual loss \eqref{eq:loss-Q-individual}, global loss \eqref{eq:loss-Q-global}, and consistency regularizer \eqref{eq:regularizer} into the overall objective:
\begin{equation}\label{eq:overall}
\begin{split}
\Lcal_{tot}(w, \theta) &= \Lcal(w) + \Lcal(\theta) + \lambda \Lcal_{reg}
\end{split}
\end{equation}
where $\lambda$ controls the extent of regularization.
Whenever any agent learns to attains high individual reward at the cost of global performance, which is likely when minimizing the individual loss, the consistency loss will increase to reflect inconsistency between individual and global performance; it will then decrease once global information influences the learning of individual $Q^n$.
Our experiments provide evidence of this dynamic learning process that balances individual and global learning (\Cref{fig:consistency-loss}).
Since the third term is a regularizer, which in general is not necessarily zero at convergence, \eqref{eq:overall} does not force $Q^{\pibf}_w$ to equal the weighted sum of all $Q_{\theta}$. 
Even at equality, agents still retain cooperation and are not independent because $Q^{\pibf}_w$ is trained using the total reward and \eqref{eq:regularizer} weighs each agent by $k^n$.
Optimizing $\Lcal(w)$ by itself does not enable action selection due to combinatorial explosion of the joint action space; optimizing $\Lcal(\theta)$ alone amounts to IQL; and, crucially, removing our novel regularization term from \eqref{eq:overall} would decouple the global and individual losses and reduce \eqref{eq:overall} to IQL.

At each training step, we interleave the updates to $\theta$ and $w$ to exchange information between the global and individual parts, allowing each agent to learn a policy that considers the effects of other agents' learning.
The parameters $w$ and $\theta$ are updated by gradient descent (derived in \Cref{app:gradients}):
\begin{equation}
\begin{split}
&\nabla_{w}\mathcal{L}_{tot} =
-E_{\pibf} \Bigl[ \bigl[ Y -(1 + \lambda)Q^{\pibf}_w(s_t, \abf_t) \\
&+ \lambda \sum_{n}k^{n}Q^{n}_{\theta}(o^n_t,a^n_t) \bigr] \nabla_{w} Q^{\pibf}_w(s_t, \abf_t) \Bigr] \\
&Y := R^g + \gamma Q^{\pibf}_{\hat{w}}(s', \abf')\vert_{\abf' = \lbrace \argmax_{a^n} Q_{\hat{\theta}}(o'^n,a^n) \rbrace}
\end{split}
\end{equation}

\begin{equation}
\begin{split}
& \nabla_{\theta}\mathcal{L}_{tot} = 
-\Ebb_{\pibf} \Bigl[ \sum_{n=1}^N \Bigl( Z_1 + \lambda k^n Z_2 \Bigr) \nabla_{\theta} Q^n_{\theta}(o^n_t,a^n_t) \Bigr] \\
&Z_1 := \frac{1}{N} \left( R^n_t + \gamma \max_{\ahat^n} Q^n_{\hat{\theta}}(o'^n,\ahat^n) - Q^n_{\theta}(o^n_t,a^n_t) \right) \\
&Z_2 := Q^{\pibf}_w(s_t,\abf_t) - \sum_{m=1}^N k^m Q^m_{\theta}(o^m_t,a^m_t)
\end{split}
\end{equation}

\section{Experimental Setup}
We evaluated the performance of our method against a large set of baselines (described in \Cref{app:algs}) on multiple road networks under a variety of traffic conditions in the SUMO simulator \cite{wu2017flow,SUMO2018}.
We describe all key experimental setup details in this section.
\Cref{sec:results} provides detailed analysis of each algorithm's performance. 
For each algorithm, we report the mean of five independent runs, with standard deviation reported in \Cref{table_conditions}.

\subsection{Environment}
\begin{figure}
\begin{minipage}[c]{0.44\textwidth}
\minipage{0.4\textwidth}
\subfigure[][]{
\label{fig:intersection}
\includegraphics[width=\linewidth,height=1.2in]{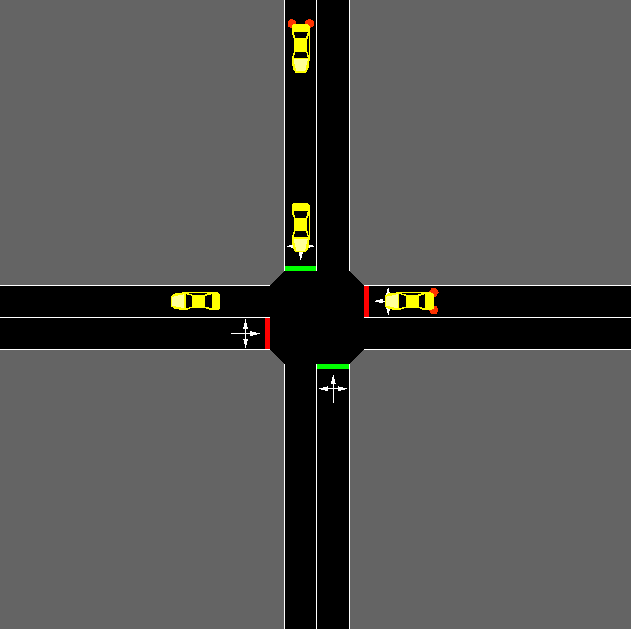}}
\endminipage\hfill
\minipage{0.4\textwidth}
\subfigure[][]{
\label{fig:1by2}
\includegraphics[width=\linewidth,height=1.2in]{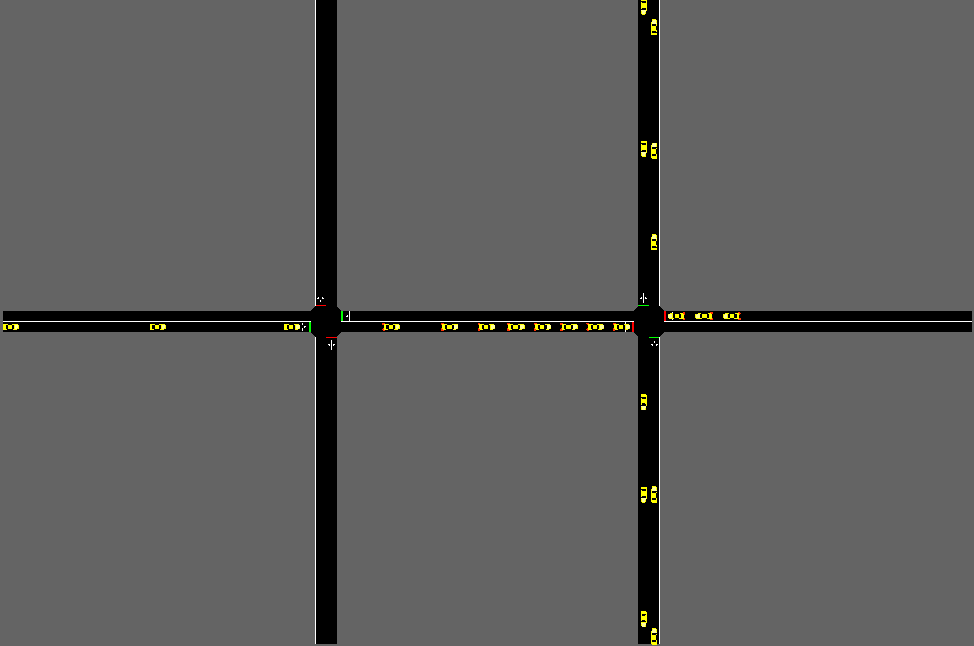}}
\endminipage\hspace{0.1cm}
\vfill
\minipage{0.4\textwidth}
\subfigure[][]{
\label{fig:2by2}
\includegraphics[width=\linewidth,height=1.2in]{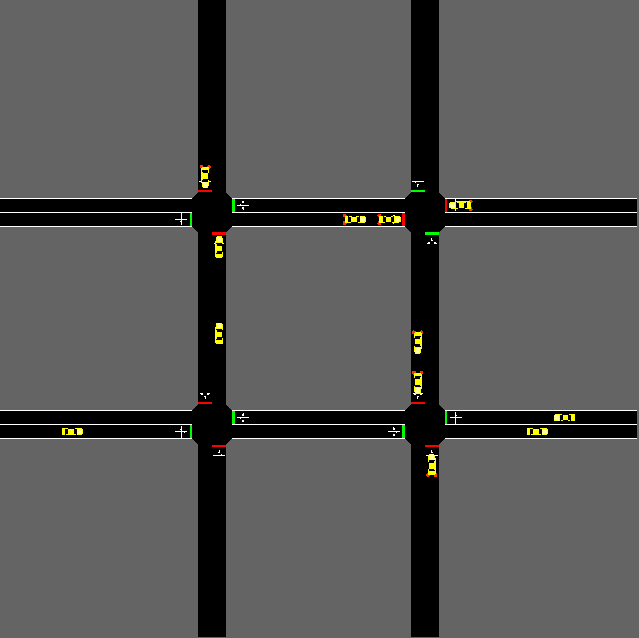}}
\endminipage\hfill
\minipage{0.4\textwidth}
\subfigure[][]{
\label{fig:6by6}
\includegraphics[width=\linewidth,height=1.2in]{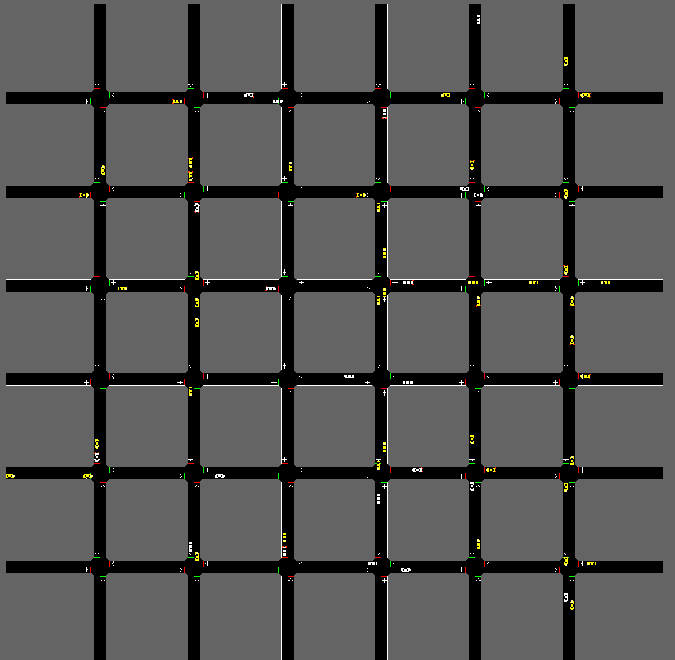}} 
\endminipage\hspace{0.05cm}
\caption[A set of three subfigures.]{Grid topology used :
\subref{fig:intersection} 1 traffic light example;
\subref{fig:1by2} 2 traffic lights;
\subref{fig:2by2} $2\times2$ traffic lights;
\subref{fig:6by6} $6\times6$ traffic lights;}%
\label{Fig:duel}%
\end{minipage}
\hfill
\begin{minipage}[c]{0.53\textwidth}%
\begin{tabular}[c]{@{}c@{}|@{}c@{}|@{}c@{}@{}c@{}}
\hline
Program&Time Period& \multicolumn{2}{c}{Num Vehicles/Hour}\\
And Topology&A:train&Horizontal&Vertical\\
&B:test&(Bot to Top)&(Left to Right)\\\hline
$1\times2$ two Agents & A:0-12000                & 700 &                        10, 620                                                                              \\ \hline
$2\times2$ Same State & A:0-12000                 & 700, 700                                                                            &  700, 700                                                                             \\ \hline
Generalization    & A:0-12000                & 700, 280                                                           &  10, 620                                                                              \\
different flow 1  & B:0-2000               & 700, 280                                                                           &  10, 620                                                                              \\
 $2\times2$  & B:2000-4000        & 1000, 800                                                                          &  900, 700                                                                             \\
                  & B:4000-6000        & 1400, 1000                                              &  400, 900                                                                             \\ \hline
Generalization    & A:0-12000                  & 1000, 580                                                                           &  110, 920                                                                             \\
different flow 2  & B:0-2000                & 1000, 580                                                                           &  110, 920                                                                             \\$2\times2$
                  & B:2000-4000              & 1000, 800                                                                           &  900, 700                                                                             \\
                  & B:4000-6000              & 1400, 1000                         &  400, 900                                                                             \\ \hline
Generalization    & A:0-12000                 & 700, 700                                                                            &  700, 700                                                                             \\
different flow 3  & B:0-2000              & 700, 700                                                                            &  700, 700                                                                             \\
     $2\times2$            & B:2000-4000                    & 1000, 800                                                                           &  900, 700                                                                             \\
                  & B:4000-6000               & 1400, 1000                                                                          &  400, 900                                                                             \\ \hline
Train on $2\times2$      & A:0-12000                & 700, 700                                                                            &  700, 700                                                                             \\
test on $6\times6$       & B:0-12000               & 700,280,                                                         &  10,620,                                                \\
& &260,240,&620,50,\\
& &780,200&90,700\\
\hline
\end{tabular}\\
{\captionof{table}{Network topologies, flow schedules and flow \\rates in all training and testing configurations}
  \label{table:config}}
\end{minipage}
\vspace{-7pt}
\end{figure}

We used the Flow framework \citet{wu2017flow} with the SUMO simulator.
We used homogeneous vehicles of the same type and size in all experiments.
Extension to heterogeneous vehicles requires no modification to our algorithm, but only redefinition of observation vectors (e.g., a longer vehicle takes up two units in queue length).
Road networks are defined as the intersection of $m$ horizontal and $n$ vertical roads (e.g. a $1\times2$ network has one horizontal route intersecting two vertical routes.). 
Each traffic light situated at an intersection is a learning agent.
Each road between two intersections is 400m long and has two lanes with opposite directions of travel.
Hence each agent has one incoming and one outgoing lane for each edge. 
Vehicles are emitted at the global outer boundaries of each edge with random starting lane and fixed entering speed. 
We used different traffic flow programs which vary in the number of vehicles per hour in the specific period to show our findings.
\Cref{table:config} contains all traffic configurations. 
At each time step, the traffic light is green exclusively for either the horizontal or the vertical direction.

\textbf{Uniform flow on symmetric $2\times2$ network.}
The first road network is a symmetric environment with $N=4$ traffic light agents defined by a $2\times2$ network \Cref{fig:2by2}.
Vehicle emissions from all route boundaries are equal and uniform throughout the entire training horizon.
Due to symmetry, all agents' rewards are weighted equally, and all agents should learn the same policy or value functions.

\textbf{Non-uniform traffic flow on $1\times2$ network.}
The second traffic control scenario has more challenging traffic dynamics. 
One horizontal route and two vertical routes form two interactions (\Cref{fig:1by2}). 
The horizontal route has 700 vehicles/hour, one vertical route on the left has only 10 vehicles/hour, while the second vertical route on the right has 600 per hour.
This is a common real-world scenario where one main arterial road with dense traffic is adjacent and parallel to a smaller local road with sparse traffic.
The greedy strategy for the left traffic light is to let vehicles in the E-W direction pass almost all the time, which would lead to high incoming E-W traffic for the second traffic light that already has heavy N-S traffic.
Hence cooperative learning is necessary for the left light to close the E-W route to some extent, to optimize global performance.

\textbf{Generalization To Different Flows.}
We investigated how well policies learned by each algorithm in one traffic condition generalize to different traffic conditions without further training. 
This is crucial for real-world applicability since training on every possible traffic condition is not feasible.
In the first experiment, we trained a policy under a \textit{static} traffic flow in the $2\times2$ network, but tested it on three consecutive equal-duration time periods with \textit{different vehicle densities}.
Denoting the flow as a vector $f:=(\text{bot, up, left, right}) \in \Rbb^4$ specifying the number of vehicles per hour on each vertical and horizontal route, traffic flows in the second and third test periods are $f_{t_2}:= (1000,800,900,700)$ $f_{t_3}:=(1400,1000,400,900)$ (\Cref{table:config} third row).
The first test period has the same traffic flow as the training phase.

We further investigated the extent to which generalization performance is affected by the specific traffic condition used in training. 
Specifically, we trained separate QCOMBO policies with \textit{different flows} in the same network topology (with one policy per training flow), and test all policies under the \textit{same} flow. 
We denote the $i$-th train-test combination ($i = 1,2,3$) as 4-component sequence $F^i = \lbrace f^i_{t_0},f^i_{t_1},f_{t_2},f_{t_3} \rbrace$, where the first flow is the training flow followed by three test flows, and $f_{t_2}$ and $f_{t_3}$ are shared by all train-test combinations.
Then the three train-test programs are: 
$F^1= \lbrace f^1_{t_0},f^1_{t_1}, f_{t_2},f_{t_3} \rbrace$, 
$F^2=\lbrace f^2_{t_0}, f^2_{t_1},f_{t_2},f_{t_3}\rbrace$,
$F^3=\lbrace f^3_{t_0},f^3_{t_1},f_{t_2},f_{t_3}\rbrace$,
where $f^1_{t_0}= f^1_{t_1}:=(700,280,10,620)$, $f^2_{t_0}= f^2_{t_1}:=(1000,580,110,920)$, $f^3_{t_0}= f^3_{t_1}:=(700,700,700,700)$ are training flows and the testing flow in the first $1/3$ period. 
Three programs use the same testing flow for the second and third periods ($f_{t_2}, f_{t_3}$), specified in \Cref{table:config}. 
Time periods $t_0,t_1,t_2,t_3$ are 10000,1000,1000,1000 steps, respectively.

\textbf{Generalization to larger networks.}
Directly training on simulations of large real-world traffic networks may not be computationally practical due to the combinatorially-large state and joint-action spaces, regardless of independent or centralized training.
Previous approaches first formulated simple models for regional traffic and then realigned to the whole system \citet{esser2000large}, or relied on transfer planning \citet{van2016coordinated}.
In contrast, we investigated the feasibility of training on a sub-network and directly transferring the learned policies without further training into a larger network.
Direct deployment in a larger systems is possible as QCOMBO learns decentralized policies.
We constructed a $6\times6$ traffic network with 36 traffic lights (\Cref{fig:6by6}) and nonuniform traffic flows, which severely reduces the possibility that any traditional hand-designed traffic control plan can be the optimal policy.
Policies trained via QCOMBO in the $2\times2$ network were directly tested on the $6\times6$ network.

\subsection{Algorithm implementations}

We implemented all algorithms using deep neural networks as function approximators.
We ensure that all policy, value, and action-value functions have the same neural network architecture among all algorithms---to the extent allowed by each algorithm (e.g. hypernetwork architecture required by QMIX)---for fair comparison.
The individual utility functions $Q^n$ of IDQN, VDN, QMIX, and QCOMBO are represented by fully-connected three-layer neural networks, where the last layer has $|A|$ output nodes.
QMIX has a two level hypernetwork that generates the weight matrix and bias vector from the global state $s$, to compute the inner product with each $Q^n$ and produce one state action value $Q(s,\abf)$.
VDN takes the sum of $Q^n$ to calculate the total $Q$, which is minimized by the squared error loss.
The critic of IAC is a value function $V$, which is used to estimate the TD error. 
The COMA uses a centralized $Q$ minus a counterfactual baseline to compute the COMA policy gradient. 
The critics of COMA and IAC have the same three-layer neural network structure, similar to the Q-functions in IDQN, VDN, QMIX. 
Both IAC and COMA use the same actor network to approximate $\pi(a_t|o_t)$, which is also a three-layer fully connected neural network that takes the agent's observation, with a softmax activation in the last layer. 
We give the agent's observations, last actions, and one-hot vector of agent labels as input to utility functions ($Q^n$ of IDQN, QMIX, and VDN, $V^n$ of IAC)
We give the global state, all other agents' actions, and agent labels as inputs to $Q$ of COMA. 
Motivated by the possibility of periodic behavior of traffic, we also experimented with RNN and GRU cells for $Q^n$ of IDQN, VDN, QMIX and QCOMBO, and the $\pi(a_t|o_t)$ of IAC and COMA.
\Cref{app:architecture} contains more architecture details.

\section{Results}
\label{sec:results}

\begin{figure*}[t]
\hspace*{0.7cm}%
\minipage[c]{0.4\textwidth}
\includegraphics[scale=0.7]{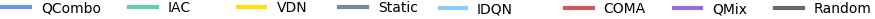}
\endminipage\vfill
\minipage{0.25\textwidth}
\subfigure[][$2\times2$]{%
\label{fig:reward_four_agents}%
\includegraphics[width=\linewidth,height=1.2in]{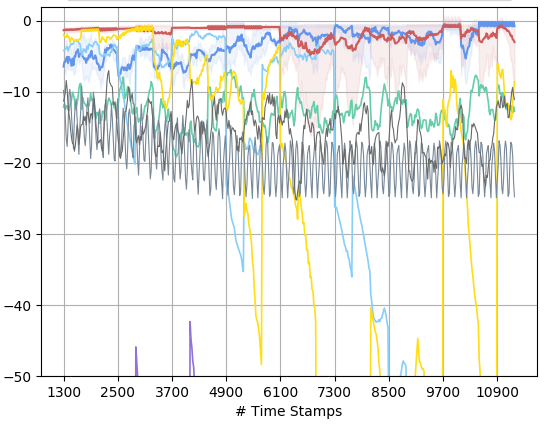}}%
\endminipage\hfill%
\minipage{0.25\textwidth}
\subfigure[][2 traffic lights]{%
\label{fig:reward_two_agents}%
\includegraphics[width=\linewidth,height=1.2in]{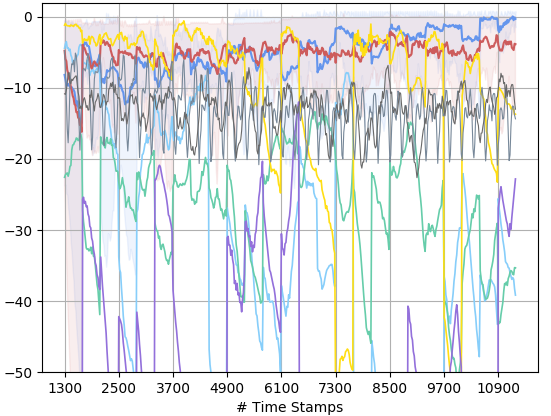}}%
\endminipage\hfill%
\minipage{0.25\textwidth}
\subfigure[][$2\times2$ using RNN]{%
\label{fig:reward_four_agents_rnn}%
\includegraphics[width=\linewidth,height=1.2in]{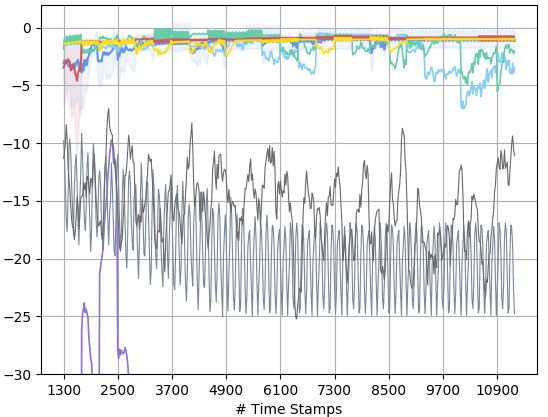}}%
\endminipage\hfill%
\minipage{0.25\textwidth}
\subfigure[][2 lights using RNN]{%
\label{fig:reward_two_agents_rnn}%
\includegraphics[width=\linewidth,height=1.2in]{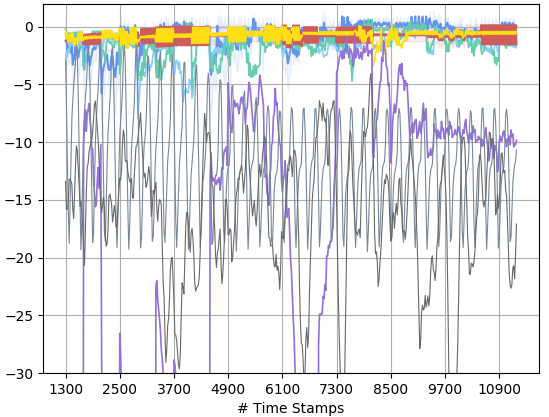}} %
\endminipage\hfill
\vspace{-10pt}
\caption[]{
\subref{fig:reward_four_agents}-\subref{fig:reward_two_agents_rnn} show rewards under different topology among algorithms}%
\label{fig:rewards}%
\end{figure*}

To analyze differences in algorithms, we show full learning curves for all algorithms in addition to reporting final performance.
This is crucial because previous work on deep RL for traffic signal control noted the potential for instability during training \cite{van2016coordinated}.
Our learning curves were generated by conducting evaluation measurements periodically during training (i.e. test the current policy for 400 steps).
Since it takes time to populate the road networks, our learning curves start after 1000 simulation time steps.
For every experiment, we ran a static policy (change light phase every 30s), and a random policy (keep or change the current phase every 5s) so that improvements due to learning can be clearly seen.

\begin{table*}[t]
\centering
\captionof{table}{ Final Traffic Condition After Learning}
\label{table_conditions}    
\begin{tabular}{c|@{}c@{}@{}c@{}@{}c@{}@{}c@{}@{}c@{}@{}c@{}@{}c@{}@{}c@{}@{}c@{}@{}c@{}@{}c@{}@{}c@{}}
\hline
\multicolumn{1}{l}{} & \multicolumn{3}{c}{$2\times2$ balanced}                                                                                & \multicolumn{3}{c}{$1\times2$ unbalanced}                                                                                \\ \cline{2-7} 
\multicolumn{1}{l}{} & Queue Length& Wait Time& Vehicle Delay &   Queue Length& Wait Time& Vehicle Delay\\ \hline
QCOMBO                  & \textbf{1.80}(0.05)       & \textbf{0.03}(0.00)      & 3.30(0.01)       & \textbf{1.12} (0.96)       & \textbf{0.14}(0.20)      & \textbf{2.08}        (0.11)       \\
IDQN                 & 23.92(24.98)      & 191.98(270.65)    & 3.44(0.29)       & 29.78(22.97)      & 44.63(52.58)     & 2.37   (0.95)       \\
IAC                  & 41.89(1.71)       & 9.16 (1.29)      & 3.45(0.02)       & 34.37 (33.20)      & 42.13 (78.92)     & 2.27 (1.14)       \\
COMA                 & 2.10(0.30)       & 0.05 (0.01)      & \textbf{3.26}(0.03)       & 7.72(8.53)       & 0.84            (1.06)      & 2.19(1.25)       \\
QMIX                 & 25.10(32.10)      & 115.72(163.57)    & 3.57(0.31)       & 16.91(19.11)      & 40.57 (58.34)     & 2.82 (0.12)       \\
VDN                  & 38.12(29.13)      & 106.34(105.83)    & 3.45(0.30)       & 10.29(15.08)      & 16.13 (34.09)     & 2.91 (0.47)       \\
Random               & 26.57(0.00)       & 2.49(0.00)      & 3.65 (0.00)       & 25.15(7.77)       & 4.96(1.37)      & 2.81(0.49)       \\
Static               & 36.14(0.00)       & 6.90(0.00)      & 3.51 (0.00)       & 27.59(4.51)       & 4.18(1.60)      & 2.83(0.02)       \\ \hline
\end{tabular}
\vspace{-10pt}
\end{table*}


\subsection{Static Traffic Flows}
\Cref{fig:rewards} shows the global reward in both the $2\times2$ and $1\times2$ road networks using both fully-connected and RNN neural networks.
Over all flow and network configurations, QCOMBO attained the global optimal performance and is most stable among all algorithms.
Policy-based methods COMA and IAC, which mitigate issues with state aliasing and are more robust to small changes of observation,
show lower variance and higher average reward than other value-based methods (IDQN, VDN and QMIX).

In the $2\times2$ environment \Cref{fig:reward_four_agents}, QCOMBO and COMA converged to global optimal policies. 
Both two maintain good traffic condition throughout learning, as reflected by queue length, waiting time, and delay time of vehicles in \Cref{table_conditions}. 
VDN performed better than the random and static policies, but worse than QCOMBO and COMA, as it can only learn a restricted set of linearly-combined Q functions.
IDQN and IAC were not stable and failed to reach a global optimum, showing that learning cooperation from the global reward is necessary.
QMIX diverged possibly due to the difficulty of stabilizing its hypernetwork.
All algorithms (with the exception of QMIX) improved with the use of RNN for policy or value functions (\Cref{fig:reward_four_agents_rnn}), giving strong evidence that RNNs are especially suitable for handling history information and periodicity of traffic.

Performance differences between algorithms are more apparent in the $1\times2$ environment with non-uniform traffic flow (\Cref{fig:reward_two_agents}).
QCOMBO converged to the optimal policy, exceeding the performance of all other algorithms.
VDN began with high performance but struggled to maintain a good policy as more vehicles enter the road.
QCOMBO's higher performance and stability over IDQN shows that the new regularization loss in QCOMBO helps to stabilize learning of independent utility functions by limiting their deviation from the centralized action-value function.
The benefit of centralized learning (e.g. QCOMBO and COMA) over independent learning (e.g. IDQN, IDQN) is more apparent than in the $2\times2$ environment, since the non-uniform flow increases the impact of each agent on other agents' performance, resulting in the need for cooperation.  
IDQN, VDN and IAC exhibits oscillation, similar to behavior reported in \cite{van2016coordinated}.
QMIX suffers from convergence issues when using a fully-connected network (RNN results shown in \Cref{fig:reward_two_agents_rnn}).

\begin{figure*}[t]
\minipage{0.16\textwidth}
\label{fig:policy_qcombo}%
\includegraphics[scale=0.25]{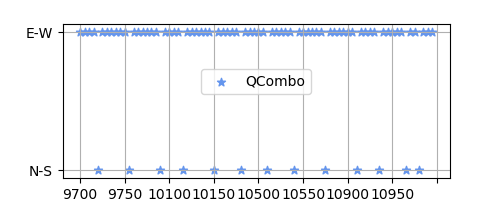}
\endminipage\hfill
\minipage{0.16\textwidth}
\label{fig:policy_coma}%
\includegraphics[scale=0.25]{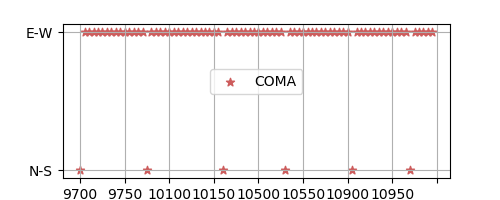}
\endminipage\hfill
\minipage{0.16\textwidth}
\label{fig:policy_iqdn}%
\includegraphics[scale=0.25]{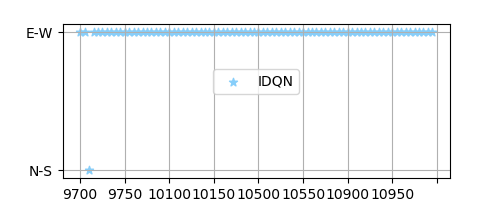}
\endminipage\hfill
\minipage{0.16\textwidth}
\label{fig:policy_iac}%
\includegraphics[scale=0.25]{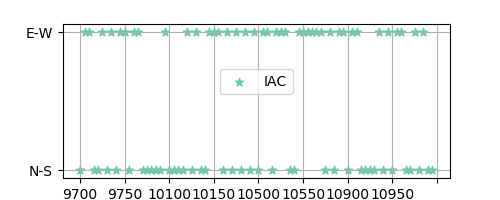}
\endminipage\hfill
\minipage{0.16\textwidth}
\label{fig:policy_vdn}%
\includegraphics[scale=0.25]{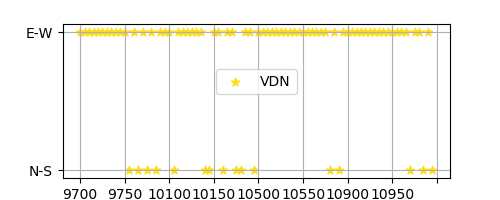}
\endminipage\hfill
\minipage{0.16\textwidth}
\label{fig:policy_qmix}%
\includegraphics[scale=0.25]{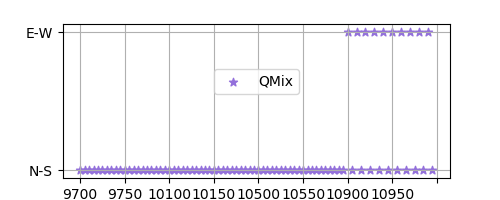}
\endminipage\hfill
\vspace{-7pt}
\caption[A set of six subfigures.]{Traffic light phase selected by the left agent in the $1\times2$ topology}
\label{fig:phases}%
\end{figure*}

We further explain differences in algorithm performance in $1 \times 2$ by analyzing actions (E-W or N-S phase) selected by the learned policies.
In order to achieve cooperation, the left traffic light should open N-S and close E-W periodically to reduce incoming E-W traffic for the right light, who already experiences heavy N-S traffic. 
\Cref{fig:phases} shows that QCOMBO and COMA achieve cooperation by turning off E-W traffic periodically (with low frequency, since it receives higher E-W than N-S traffic).
However, IDQN greedily keeps E-W open for long durations, which is not globally optimal; IAC switches between the two phases almost equally; VDN switches to N-S too often than necessary; QMIX incorrectly chooses N-S more frequently than E-S. 

\subsection{Generalizing To Dynamic Traffic Flows}

\begin{figure}[t]
\centering
\minipage{0.33\textwidth}
\subfigure[][]{
\label{fig:algos_flows}
\includegraphics[width=\linewidth,height=1.4in]{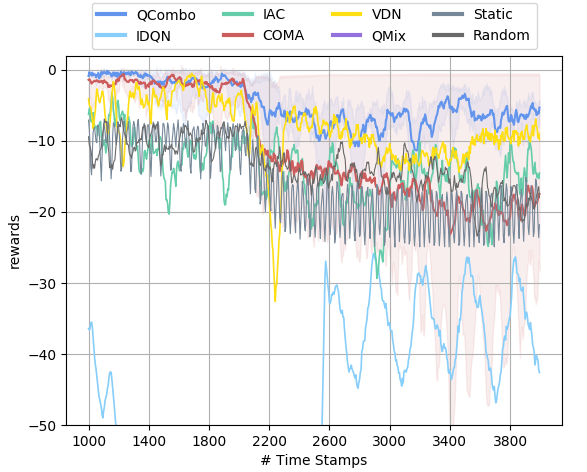}}
\endminipage\hfill
\minipage{0.33\textwidth}
\subfigure[][]{\label{fig:QCOMB_flows}
\includegraphics[width=\linewidth,height=1.4in]{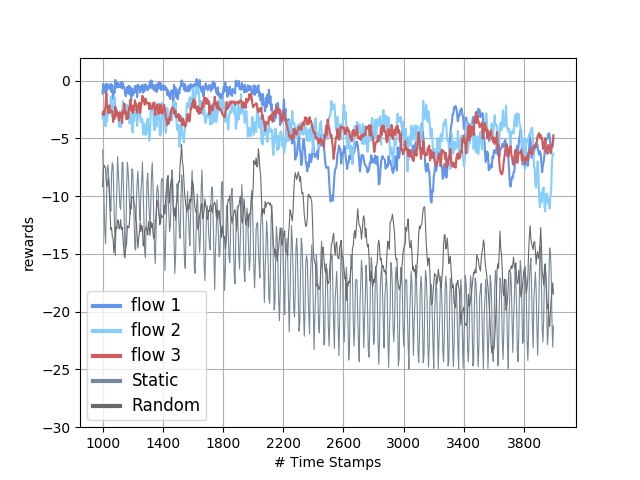}}
\endminipage\hfill
\minipage{0.33\textwidth}
\subfigure[][]{\label{fig:largesystem}
\includegraphics[width=\linewidth,height=1.4in]{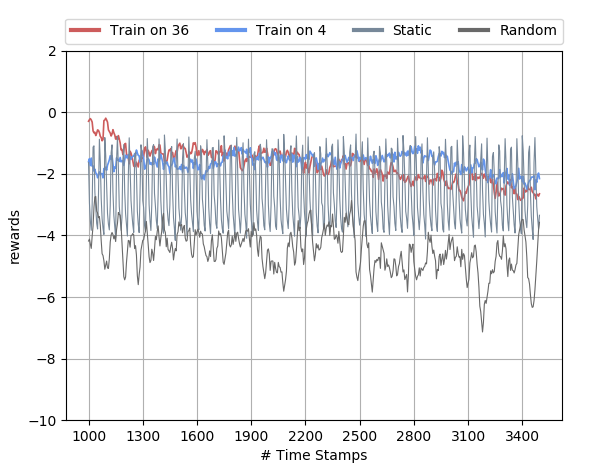}} 
\endminipage
\vspace{-10pt}
\caption[]{More Generalization Results,
\subref{fig:algos_flows} generalization to different traffic flows among algorithms;
\subref{fig:QCOMB_flows} Impact of training conditions on performance in new test condition; \subref{fig:largesystem} Policy trained in $2\times2$ network generalizes to $6\times6$ network}%
\label{fig:generalization}%
\end{figure}

QCOMBO displayed the highest generalization performance, when trained on one traffic condition and deployed on two different test conditions \Cref{fig:algos_flows}.
As reflected by the decrease in performance of all policies when traffic flow changes at time step 2000, the test conditions were more difficult.
While COMA and QCOMBO have equal performance on the training flow (first 1000 steps), QCOMBO generalized much more gracefully to the test conditions as the consistency regularizer prevents overfitting to local training conditions.
VDN perform worse on the training flow than COMA but generalized better, despite experiencing a large drop in the second test flow when the vehicle density increases.
IAC shows high variance on the test condition, while IDQN and QMIX could not adapt to new flows due to low training performance.

\Cref{fig:QCOMB_flows} shows results of the second generalization experiment, where we evaluated QCOMBO policies trained on three different traffic flows, using the same test program. 
The first 1000 steps have the same flow as during training, while the new flows appear at $t=2000$ and $t=3000$.
QCOMBO policies show flow invariance during the $t_2$ and $t_3$ period: trained on different traffic conditions, generalization performance on new unseen conditions exhibit only small variability.
This gives evidence that performance of QCOMBO on test conditions does not heavily depend on specific choices of training conditions.

\subsection{Generalization To Larger Traffic Topology}

We directly applied the QCOMBO policy trained in the $2\times2$ traffic network to the $6\times6$ network with 36 traffic light agents, which poses a significant generalization challenge due to the decreased observability for any particular agent and the different traffic flow induced by the different network topology.
\Cref{fig:largesystem} shows that QCOMBO's policy is able to maintain high and stable test performance, with almost negligible difference from its training performance.
Surprisingly, it sometimes attains higher reward even than a policy that was trained specifically on the $6\times6$ environment.
This shows that centralized training with few agents can still produce policies that generalize well to larger settings, mitigating concerns about scalability of centralized training.
This is strong evidence that a policy trained on a subset of a city road network can be deployed with little loss of performance at a larger scale.

\subsection{Conclusion}
We proposed QCOMBO, a novel MARL algorithm for traffic light control. 
QCOMBO combines the benefits of independent and centralized training, with a novel objective function that enforces consistency between a global action-value function and the weighted sum of individual optimal utility functions.
We conducted detailed empirical evaluation of state-of-the-art MARL algorithms (IDQN, IAC, VDN, COMA, and QMIX) on network traffic light control under different map topologies and traffic flows, and showed that QCOMBO is a simple yet competitive approach to this real-world problem.
Experiments also indicate that QCOMBO can be generalized with limited loss of performance to large traffic networks.
Our work gives strong evidence for the feasibility of training cooperative policies for generalizable, scalable and intelligent traffic light control.

\bibliographystyle{natbib}
\bibliography{main}

\newpage 
\appendix
\appendixtitleon
\appendixtitletocon

\begin{appendices}




\section{Environment and Experiment}
\subsection{SUMO Setup}
One time step in SUMO corresponds to 0.1s in the real world. We limit the traffic light to control vehicles going straight only. Table \ref{table_SUMO} has the configuration of SUMO environment.   

\begin{table*}[ht]
  \centering
\caption{The SUMO Configuration}\label{table_SUMO}
\scalebox{0.85}{
\begin{tabular}{llll}
\hline
             & Parameter          & Value              & Description                                                                                                                                                                                                                                                                                                 \\ \cline{2-4} 
vehicle      & maxSpeed           & 35m/s              & The maxium alllowed speed in the lane                                                                                                                                                                                                                                                                       \\
             & minGap             & 2m                 & The minimum empty space that vehicles are allowed to have                                                                                                                                                                                                                                                                                  \\
             & tau                & 1s                 & \begin{tabular}[c]{@{}l@{}}The minimum time gap of tau between the rear bumper \\ of vehicle's leader and its own front-bumper + minGap \\ to assure the possibility to brake in time when its leader \\ starts braking and it needs $tau$ seconds reaction time to start breaking as well\end{tabular} \\
             & spacing            & uniform            & The positioning of vehicles in the network relative to one another                                                                                                                                                                                                                                          \\
             & accel              & $1.0m/s^2$      & The acceleration ability                                                                                                                                                                                                                                                                                    \\
             & decel              & $1.5m/s^2$     & The deceleration ability                                                                                                                                                                                                                                                                                                                                                                                                                                                                                                                                                                        \\
             & speed\_factor      & 1                  & The vehicles expected multiplicator for lane speed limits                                                                                                                                                                                                                                                   \\
             & speed\_dev         & 0.1                & The deviation of the speedFactor                                                                                                                                                                                                                                                                                \\
             & car\_follow\_model & IDM                & \begin{tabular}[c]{@{}l@{}}The numerical integration method used to control the dynamic update of the \\ simulation,  the intelligent driver model results in very conservative lane \\changing gap acceptance.\end{tabular}                                                                                    \\ \hline
traffic light & phase              & green, yellow, red & The traffic light alternating phase                                                                                                                                                                                                                                                                         \\
             & green              & 30s                & The green light duration                                                                                                                                                                                                                                                                                    \\
             & yellow             & 3s                 & The yellow light duration                                                                                                                                                                                                                                                                                   \\
             & red                & 30s                & The red light duration                                                                                                                                                                                                                                                                                      \\
             & tls\_type          & static             & \begin{tabular}[c]{@{}l@{}}Static traffic lights are traffic lights with pre-defined phases, \\ they cannot dynamically adjust according to traffic needs; \\ they simply follow the same pattern repeatedly.\end{tabular}                                                                                  \\ \hline
\end{tabular}}

\end{table*}

\subsection{Reward Definition}
\label{app:reward-definition}
The individual reward $R^n = c_1\times ql^n + c_2\times wtl^n + c_3 \times dl^n + c_4 \times eml^n + c_5\times fl^n + c_6\times vl^n$ is a weighted linear combination of features $ql^n, wtl^n, dl^n, eml^n, fl^n, vl^n$ with weights $c_1,c_2,c_3,c_4,c_5,c_6$, given for each $\Delta_t$=5s interval.
\begin{itemize}[leftmargin=*]
    \item $ql^n := \sum_{\Delta_t}\sum_{l}q_l^n$ is the sum of queue length of all incoming lanes, weighted by $c_1=-0.5$
    \item $wtl^n := \sum_{\Delta_t}\sum_{l}wt_l^n$ is the sum of vehicle waiting time among all incoming lanes, with $c_2=-0.5$.
    \item $dl^n := \sum_{\Delta_t}\sum_{l}delay_l^n$ is the sum of vehicle delay of all incoming lanes queue, with $c_3=-0.5$.
    \item $eml^n$, the number of vehicles that have emergence stops on the traffic light lanes over $\Delta_t$, a vehicle is counted in this category if its current speed more than 4.5m/s less than the 1 second before, and $c_4=-0.25$
    \item $fl^n$, the number of times that traffic lights change its phase over $\Delta_t$, and $c_5=-1$.
    \item $vl^n$, the number of vehicles that pass the traffic light over $\Delta_t$, and $c_6=1$. 
\end{itemize}

\section{MARL algorithms}
\label{app:algs}

\subsection{QCOMBO Loss Curve}
\begin{figure*}[t]%
\centering
\begin{minipage}[t]{0.33\textwidth}\subfigure[][]{%
\label{fig:consistency-loss}%
\includegraphics[scale=0.2]{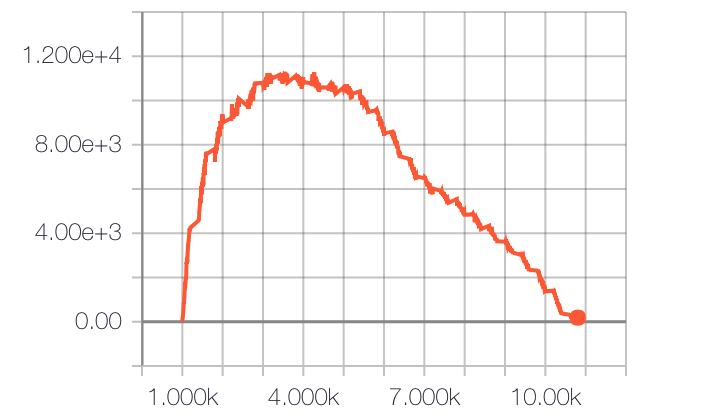}}%
\hspace{1pt}%
\end{minipage}\hfill%
\begin{minipage}[t]{0.33\textwidth}
\subfigure[][]{%
\label{fig:global-loss}%
\includegraphics[scale=0.2]{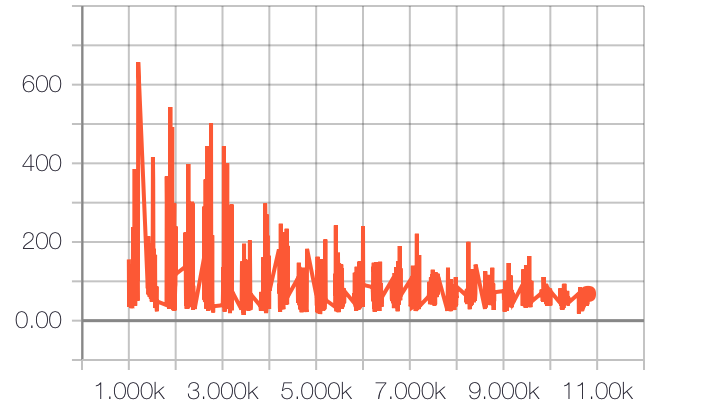}}
\hspace{1pt}%
\end{minipage}\hfill
\begin{minipage}[t]{0.33\textwidth}
\subfigure[][]{%
\label{fig:individual-loss}%
\includegraphics[scale=0.2]{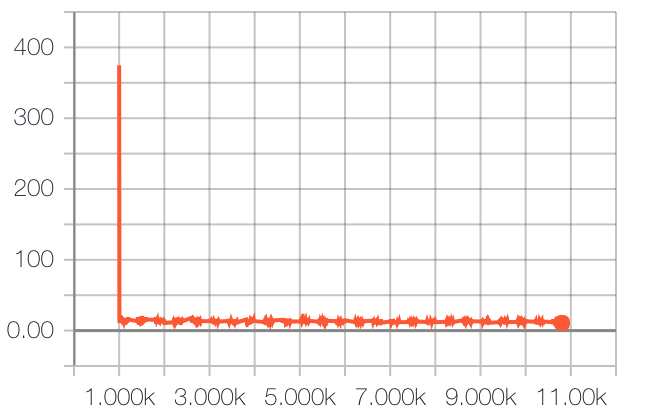}}
\end{minipage}\hfill
\caption[A set of three subfigures.]{Training Loss on $1\times2$ map:
\subref{fig:consistency-loss} consistency loss;
\subref{fig:global-loss} global loss;
\subref{fig:individual-loss} individual loss;}%
\label{fig:tensorboard}
\end{figure*}

\Cref{fig:tensorboard} shows the learning curve for loss in QCOMBO.

\subsection{IQL and IAC}

IQL directly applies single-agent Q-learning to each agent of the Markov game.
While the optimal action-value function of an MDP is defined as
\begin{align}
    Q^*(s,a) := \max_{\pi} \Ebb_{\pi} \Bigl[ \sum_{t=0}^{\infty} \gamma^t R_t \mid s_0=s, a_0=a \Bigr],
\end{align}
IQL agents learn a local utility function $Q(o^n,a^n)$ by minimizing the loss function \cite{mnih2015human}
\begin{align}
    L(\theta) = \Ebb_{\pibf} \Bigl[ \bigl( R^n_t + \gamma \max_{a^n} Q(o^n_{t+1},a^n_t;\theta') - Q(o^n_t,a^n_t;\theta) \bigr)^2 \Bigr]
\end{align}
where $\theta$ are parameters of the function approximation and $\theta'$ are parameters of a target network.
IAC directly applies the single-agent policy gradient with a variance-reduction baseline $b(s)$ \cite{sutton2000policy}
\begin{align}
    \nabla_{\theta} J(\pi) = \Ebb_{\pibf} \Bigl[ \nabla_{\theta} \log \pi(a^n|o^n) \bigl( Q^{\pi}(o^n,a^n) - b(s) \bigr) \Bigr]
\end{align}

\subsection{COMA}
The counterfactual multi-agent policy gradient \cite{foerster2018counterfactual} is
\begin{align}
\nabla_{\theta} J(\pibf) &= \Ebb_{\pibf} \Bigl[ \sum_n \nabla_{\theta} \log \pi^n(a^n|o^n) \bigl( Q^{\pibf}(s,\abf) - b(s,a^{-n}) \bigr) \Bigr] \label{eq:coma-gradient} \\
b(s,a^{-n}) &:= \sum_{\hat{a}^n} \pi^n(\hat{a}^n|o^n) Q^{\pibf}(s, (a^{-n},\hat{a}^n))
\end{align}

\subsection{VDN}
The joint action-value function $Q^{\text{VDN}}(s,\abf) := \sum_{n=1}^N Q_n(o^n,a^n)$ in VDN \cite{sunehag2017value} is trained with
\begin{align}
    L(\theta) &= \Ebb_{\pibf} \Bigl[ \bigl( y_t - Q^{\text{VDN}}(s_t,\abf_t)   \bigr)^2  \Bigr] \\
    y_t &:= R + \gamma Q^{\text{VDN}}(s_{t+1},\abf_{t+1})\vert_{\abf_{t+1} = \lbrace \argmax_{a^n} Q(o^n_{t+1},a^n) \rbrace_n}
\end{align}
Agents act greedily with respect to their own $Q^n(o^n,a^n)$.

\subsection{QMIX}
QMIX \cite{rashid2018a} enforces the monotonicy of $Q^*(s,\abf) = F(Q_1,\dotsc,Q_N)$ by using hypernetworks, which passes the state variable through linear layers followed by absolute-activation functions, to generate the weights of a mixing network that takes in all agents' $Q_n$ and produces a joint Q value.

\section{Architecture and Training}\
\label{app:architecture}

\subsection{Fully Connected Neural Network}
For standardization, the $Q^n$ of IDQN, VDN, QMIX, QCOMBO and $V^n$ of IAC use the same nerual network architecture, which is a three-layer fully-connected neural network, with 256 units in the hidden layers and ReLu activation. 
The output has $|A|$ units to represent the agent's state action value function output. 
The input is the agent's own observation vector $o^n$, its last action $a_{t-1}^n$, and one hot agent label $n$. 
The actors of IAC and COMA use the same neural network structure, which is a three-layer fully-connected network with 64 hidden units and ReLu activation. 
The output of actor neural network uses the softmax activation, and the input is $o^n$. 
All agents share the same parameter $\theta$. 
The global $Q^{\pibf}$ of QCOMBO is approximated by a three-layer fully-connected neural network with 256 hidden units, its input is the global state $s$ and all agents' action $\abf$, and the output is a single node. 
The $Q^{\pibf}$ of COMA also has three layers, and has inputs $\{s,\abf^{-n},n,o^n\}$. 
The learning rate is 0.001 for $Q$ and $V$, and 0.0001 for the actor network. 
The target network is updated gradually in every training step as a sum of $1\%$ of online network parameters and $99\%$ of current target network parameters. 
All the network training uses the Adam optimizer.

\subsection{RNN Network}
An alternative implementation of the algorithms use RNN with GRU cell to approximate the $Q^n$ of IDQN, QMIX, VDN, QCOMBO, and the actors of COMA and IAC. 
The RNN has 64 hidden units with a Relu activation function, the input of RNN is linearly transformed from input space to 64 units, and the output of the hidden layer is linearly transformed into $|A|$ outputs to represent each discrete action. 
The hidden state is retained for each time step and also re-used for next training batch.

\subsection{Training Strategy}
All the algorithms have same training strategy. 
The total time horizon for training and online evaluation is limited to 12K SUMO time steps. 
We limit the agents to make decisions for every 5 times steps to avoid high-frequency flickering of traffic lights. 
For each training run, the environment runs freely for 1000 steps using a random policy to populate the road network with vehicles before training begins. 
Then we alternate between training and evaluation, with each training and evaluation period lasting 400 time steps. 
During every training step (once per 5 time steps), the algorithm updates parameters of the the current networks via stochastic gradient descent with 100 minibatches of 30 samples from the replay buffer. 
The replay buffer is a first-in first-out queue containing the 1000 most recent samples. 
If the RNN framework is involved, we use a different sampling strategy: 1. every training cycle contains 6 consecutive periods; 2. in each period, we collect 30 consecutive samples and feed them into the RNN in the order of their corresponding time periods from the earliest to the latest; 3. after we finish one period parameter learning with 30 time steps data, we record the hidden state and use it to initialize the hidden state for the next learning step. 
The current action is selected through an $\epsilon$-greedy behavior policy with $\epsilon$ starting with 0.9 and a decay geometric factor of 0.995. 
$\epsilon$ is decayed every training cycle and fixed within a cycle. 
A training cycle is followed with an on-policy evaluation period. 
During the evaluation period of 400 time steps, the system executes the current policy and stops decaying the $\epsilon$.
We record rewards only during the last 200 of the 400 time steps, to measure the steady-state of the system under the current policy. 

\subsection{Gradients}
\label{app:gradients}

\begin{equation}
\begin{split}
&\nabla_{w}\mathcal{L}(w,\theta) =
E_{\pibf} \Bigl[ -\bigl( y_t-Q^{\pibf}_w(s_t,\abf_t)\bigr) \nabla_{w}Q^{\pibf}_w(s_t,\abf_t)\\
&+\lambda \bigl( Q^{\pibf}_w(s_t,a_t) - \sum_n k^n Q^n_{\theta}(o^n_t,a^n_t) \bigr) \nabla_w Q^{\pibf}_w(s_t,\abf_t) \Bigr] \\
&= -E_{\pibf} \Bigl[ \bigl[ R^g + \gamma Q^{\pibf}_{\hat{w}}(s_{t+1}, \abf_{t+1})\vert_{\abf_{t+1} = \lbrace \argmax_{a^n} Q_{\hat{\theta}}(o_{t+1}^n,a^n) \rbrace_n} \\
&- (1+\lambda)Q^{\pibf}_w(s_t, \abf_t) + \lambda \sum_{n}k^{n}Q^{n}_{\theta}(o^n_t,a^n_t) \bigr] \nabla_w Q^{\pibf}_w(s_t,\abf_t) \Bigr]
\end{split}
\end{equation}

\begin{equation}
\begin{split}
& \nabla_{\theta}\mathcal{L}(w,\theta) = -\E_{\pibf} \Bigl[ \frac{1}{N} \sum_{n=1}^N  \bigl( y^n_t - Q^n_{\theta}(o^n_t, a^n_t) \bigr) \nabla_{\theta} Q^n_{\theta}(o^n,a^n)\\
&+ \lambda \Bigl( Q^{\pibf}_w(s_t,a_t)-\sum_{m=1}^N k^m Q^m_{\theta}(o^m_t, a^m_t) \Bigr) \nabla_{\theta} \sum_{n=1}^N k^n Q^n_{\theta}(o^n_t,a^n_t) \Bigr]\\
&= -\Ebb_{\pibf} \Bigl[ \sum_{n=1}^N \Bigl( \frac{1}{N} \bigl[ R^n_t + \gamma \max_{\ahat^n} Q^n_{\hat{\theta}}(o^n_{t+1},\ahat^n_t) - Q^n_{\theta}(o^n_t,a^n_t)\bigr]  \\
&+ \lambda k^n \bigl[ Q^{\pibf}_w(s_t,\abf_t) - \sum_{m=1}^N k^m Q^m_{\theta}(o^m_t,a^m_t) \bigr] \Bigr) \nabla_{\theta} Q^n_{\theta}(o^n_t,a^n_t) \Bigr]
\end{split}
\end{equation}

\end{appendices}
 
\end{document}